\documentclass[11pt]{article}

\usepackage[final]{acl}

\usepackage{times}
\usepackage{latexsym}

\usepackage[T1]{fontenc}

\usepackage[utf8]{inputenc}
\usepackage{multirow}
\usepackage{booktabs}

\usepackage{microtype}

\usepackage{inconsolata}

\usepackage{float}

\usepackage{graphicx}
\usepackage{listings}
\usepackage{xcolor}
\usepackage[breakable,skins]{tcolorbox}

\lstdefinestyle{prompt}{
  basicstyle=\small\ttfamily,
  breaklines=true,
  breakatwhitespace=false,
  columns=flexible,
  keepspaces=true,
  frame=none,
  aboveskip=4pt,
  belowskip=0pt,
}

\newtcolorbox{promptbox}[1]{
  breakable,
  enhanced,
  colback=gray!5,
  colframe=gray!55,
  fonttitle=\small\bfseries,
  title={#1},
  colbacktitle=gray!15,
  coltitle=black,
  boxrule=0.4pt,
  toprule=0.4pt,
  titlerule=0.4pt,
  left=6pt,
  right=6pt,
  top=4pt,
  bottom=4pt,
}

\title{Co-Evolving Graph and Text Memory for Training-Free Multi-Hop Question Answering}

\author{Hieu Man \\
  University of Oregon, OR, USA \\
  \texttt{hieum@uoregon.edu} \\\And
  Thien Huu Nguyen \\
  University of Oregon, OR, USA \\
  \texttt{thienn@uoregon.edu} \\}

\begin{document}
\maketitle
\begin{abstract}
Multi-hop question answering requires coordinating relational and textual evidence across reasoning steps, a combination neither a text corpus nor a knowledge graph can supply alone. Prior work often emphasizes only part of this loop: graph-augmented RAG retrieves from a pre-built or query-updated graph, KGQA systems search within topic-centered subgraphs, and memory-augmented agents maintain evolving memories without continuously reconciling graph memory with textual context. We propose Co-E, a training-free system built around synchronized bidirectional graph-text working memory. A synchronization cycle consolidates textual memory, extracts relational triples into graph memory, and injects graph facts back into the generation context. Because both memories are maintained, they shape subsequent retrieval and generation. Evaluated on six multi-hop QA benchmarks, Co-E improves over comparable training-free open-backbone baselines and is competitive with larger or trained systems.
\footnote{Codebase available at \texttt{https://github.com/hieum98/wemg}}
\end{abstract}

\section{Introduction}

Retrieval-augmented generation (RAG) has become the dominant paradigm for knowledge-intensive question answering \cite{zhao2024retrievalaugmentedgenerationaigeneratedcontent, singh2026agenticretrievalaugmentedgenerationsurvey}. Multi-hop questions, however, expose a limitation deeper than retrieval recall. Later evidence often depends on entities, relations, or constraints discovered only after earlier hops are resolved. A passage may reveal a bridge entity that should expand the graph frontier; a graph edge may reveal a constraint that should reshape the next textual query. Standard RAG pipelines cannot perform this update because they select context before reasoning unfolds. Likewise, graph-augmented methods that rely on a pre-built corpus graph or a fixed topic-centered candidate subgraph cannot fully revise their relational state during reasoning, leaving missed bridge entities, noisy edges, and incomplete paths uncorrected. The central challenge is to maintain a reasoning state in which textual evidence and relational structure revise each other as the answer is constructed.


Existing systems address parts of this problem, but usually privilege one side of the reasoning state. Corpus-graph RAG systems build a graph index before inference and retrieve from it \cite{dong2026usegraphneedsefficiently, ma2026mitigatingkgqualityissues, zhou2026understandmemorycognitivegistdriven}. KGQA systems such as ToG \cite{sun2024thinkongraphdeepresponsiblereasoning} and ToG 2.0 \cite{ma2025thinkongraph20deepfaithful} improve graph-based reasoning over knowledge bases, but rely on pre-existing topology and do not maintain paired textual working memory. Graph search methods such as MCTS-KBQA \cite{xiong2025mctskbqamontecarlotree} and ReKG-MCTS \cite{song-etal-2025-rekg} navigate a fixed frontier rather than extending it. A parallel line equips agents with evolving memories: HGMem \cite{zhou2026improvingmultistepraghypergraphbased} builds hypergraph memory from retrieved text, SubQRAG \cite{li2025subqragsubquestiondrivendynamic} accumulates extracted triples as graph memory, while SE-Search \cite{li2026sesearchselfevolvingsearchagent}, MemSearch-o1 \cite{zhang2026memsearcho1empoweringlargelanguage}, and A-MEM \cite{xu2025amemagenticmemoryllm} maintain textual or reflective memories. These systems show that maintained memory is useful; what remains less explored is how textual and graph memories should revise one another in same reasoning loop.

We argue that this coupling requires \textbf{bidirectional, within-step co-evolution of graph and text}. When a passage mentions a new entity, it should become available for graph expansion; when a triple is extracted or retrieved, it should enter the textual context for the next generation step. Without both directions, the system either accumulates unstructured text that must be reinterpreted from scratch, or traverses a graph whose useful facts are not exposed to the generator. We introduce \textbf{Co-E} (\textbf{Co-E}volving), a training-free multi-hop QA system built around synchronized graph-text working memory. Co-E maintains a shared memory with two coupled stores: question-relevant text snippets and graph memory containing KB-schema and open-vocabulary triples. At each step, Co-E retrieves from text and knowledge-base streams, filters evidence, and applies a four-operation synchronization cycle: it consolidates textual evidence, extracts and merges triples into graph memory, injects surviving graph facts back into textual memory, and re-consolidates the context. This cycle lets text expand the graph frontier while graph structure grounds subsequent generation. Because the same mechanism operates during inference, Co-E requires no fine-tuning and serves both multi-hop text-QA and KGQA settings.

Empirically, Co-E achieves strong results across six benchmarks spanning text-QA and KGQA. With a Qwen3-8B backbone and no training, Co-E reaches 72.6 EM on 2WikiMultiHopQA, 70.0 EM on Bamboogle and 74.9 Hits@1 on CWQ, outperforming comparable open-backbone baselines and competing with larger or trained systems.

\section{Related Work}

We organize prior work by how evidence is represented and updated during multi-hop reasoning.

\textbf{Corpus-based retrieval and agentic search.}
Corpus-based RAG methods decompose complex questions into sub-queries and interleave retrieval with intermediate reasoning. IRCoT \cite{trivedi-etal-2023-interleaving} alternates retrieval with chain-of-thought reasoning, while HopRAG \cite{liu2025hopragmultihopreasoninglogicaware}, ComposeRAG \cite{wu2025composeragmodularcomposablerag}, and RT-RAG \cite{shi2026reasoningtreesimprovingretrievalaugmented} improve query decomposition, structured prompting, or reasoning-tree construction. A related line trains or prompts agents to search over multiple turns, such as Search-R1 \cite{jin2025searchr1trainingllmsreason} and MR-Search \cite{xiao2026metareinforcementlearningselfreflectionagentic}. These systems make retrieval adaptive, but evidence remains primarily textual. Relational structure must be inferred from snippets rather than maintained as explicit state. Co-E complements adaptive retrieval with graph memory that makes entities and relations available to later retrieval and generation.

\textbf{Graph-augmented RAG and KG reasoning.}
Graph-augmented approaches add structured evidence to RAG, but many rely on graph structure fixed before or outside the current reasoning trajectory. Corpus-graph systems such as HippoRAG 2 and C2RAG \cite{gutiérrez2025ragmemorynonparametriccontinual, ma2026mitigatingkgqualityissues} build graph-like indices over text and retrieve from them at inference time. EA-GraphRAG \cite{dong2026usegraphneedsefficiently}, GraphAnchor \cite{liu2026graphanchoredknowledgeindexingretrievalaugmented}, PAGER \cite{li2026structuredknowledgerepresentationcontextual}, and Graph-R1 + EKA \cite{wang2026multihopreasoningearlyknowledge} improve when or how structured evidence is retrieved. KGQA systems differ in how they choose the graph frontier. Learned/subgraph systems such as SubgraphRAG \cite{li2025simpleeffectiverolesgraphs}, GNN-RAG \cite{mavromatis-karypis-2025-gnn}, and iQUEST \cite{wang-yu-2025-iquest} learn or compute query-specific subgraphs before handing evidence to the LLM. Live-KG methods such as ToG \cite{sun2024thinkongraphdeepresponsiblereasoning}, ToG 2.0 \cite{ma2025thinkongraph20deepfaithful}, PoG \cite{tan2025pathsovergraphknowledgegraphempowered}, Interactive-KBQA \cite{xiong-etal-2024-interactive}, MCTS-KBQA \cite{xiong2025mctskbqamontecarlotree}, ReKG-MCTS \cite{song-etal-2025-rekg}, and KERAG \cite{sun2025kerag} focus on traversing or quering a knowledge graph during inference. Trained graph-reasoning systems such as RoG \cite{luo2024reasoninggraphsfaithfulinterpretable}, KBQA-R1 \cite{sun2026kbqar1reinforcinglargelanguage}, and GraphWalker \cite{xu2026graphwalkeragenticknowledgegraph} improve retrieval or reasoning through supervised or reinforcement learning. These systems provide strong graph retrieval; Co-E focuses on synchronizing structured and textual working state during reasoning.

\begin{figure*}[ht]
   \centering
   \includegraphics[width=\textwidth]{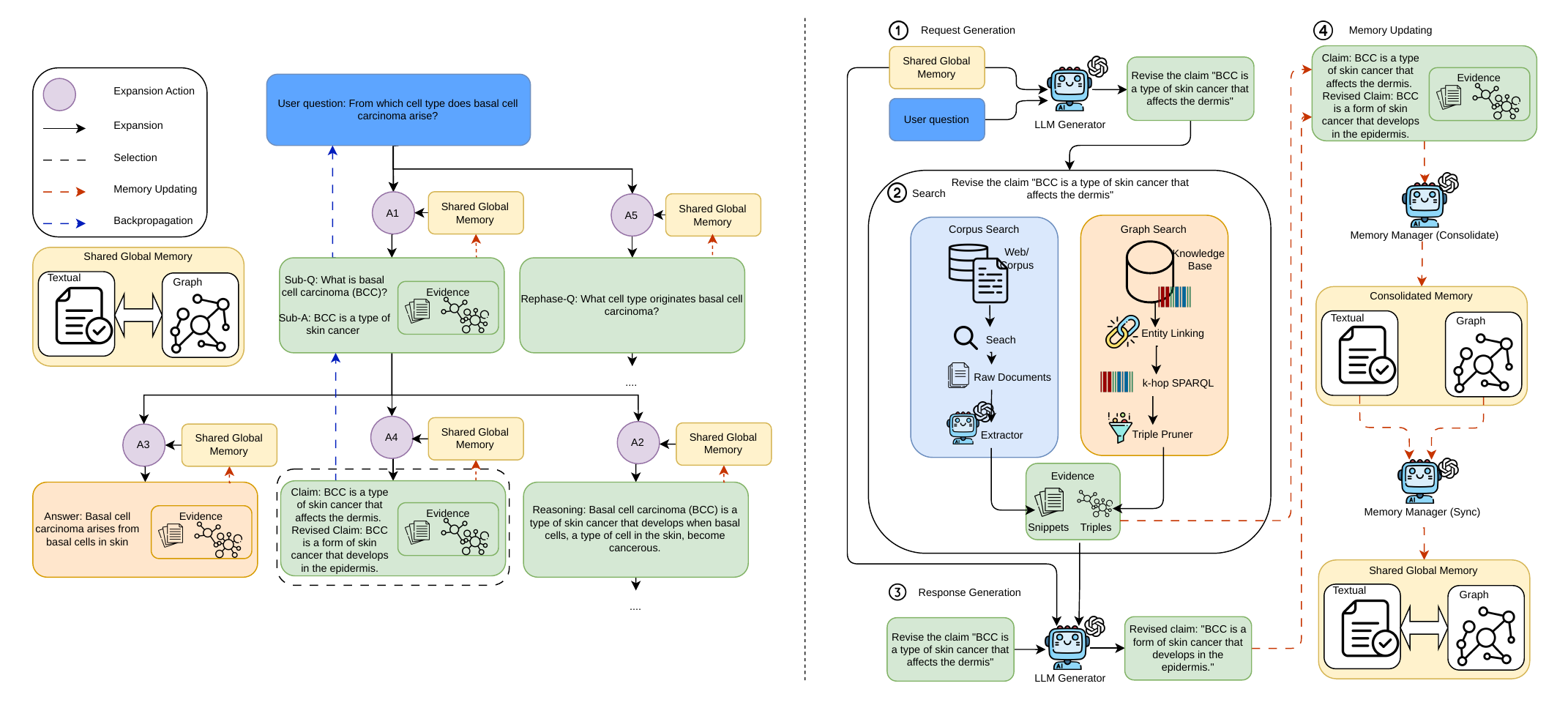}
   \caption{Co-E framework. Left: MCTS mode, where synchronized shared memory guides tree expansion. Right: the four-stage reasoning loop. Each step generates a sub-query from shared graph-text memory $\mathcal{M}^{(t)}$, retrieves text snippets $S^{(t)}$ and graph triples $R^{(t)}$, generates an intermediate answer, and applies bidirectional $\mathrm{Sync}$ to produce the next step's memory $\mathcal{M}^{(t+1)}$.}
   \label{fig:overview}
 \end{figure*}

\textbf{Memory-augmented reasoning.}
Memory-augmented reasoning systems maintain intermediate state across reasoning or search steps, whether in text, graph, or both. HGMem \cite{zhou2026improvingmultistepraghypergraphbased} builds hypergraph memory from retrieved text. SubQRAG \cite{li2025subqragsubquestiondrivendynamic} is closest to Co-E among graph-memory systems: it decomposes questions, consults source documents when the graph is insufficient, and accumulates extracted triples into graph memory. SE-Search \cite{li2026sesearchselfevolvingsearchagent} purifies evidence through a Think-Search-Memorize loop, MemSearch-o1 \cite{zhang2026memsearcho1empoweringlargelanguage} grows reasoning-aligned memory paths, and A-MEM \cite{xu2025amemagenticmemoryllm} revises textual notes retroactively. These works show that memory should be maintained, not merely accumulated. Co-E is complementary: instead of treating memory as a textual note store or graph-only trace, it maintains paired textual and graph memories and reconciles them after each reasoning step. Graph facts are therefore injected back into the textual context that conditions later retrieval and generation.

\section{Methodology}
\label{sec:method}

\subsection{System Overview}
\label{sec:overview}

Given a question $q$, Co-E answers it through an iterative reasoning process over a shared graph-text memory. At step $t$, the memory is
\begin{equation}
\mathcal{M}^{(t)} = \bigl(\mathcal{T}^{(t)}, \mathcal{G}^{(t)}\bigr),
\end{equation}
where $\mathcal{T}^{(t)}$ is textual memory and $\mathcal{G}^{(t)}$ is graph memory. Co-E produces a final answer $a$ while updating both memories after each reasoning step. Unlike pipelines that treat retrieved graph context as fixed, Co-E actively maintains $\mathcal{G}$: it can add text-derived triples, prune irrelevant edges, and write surviving facts back into textual memory.

Each reasoning step has four stages. First, \textbf{Request Generation} produces a focused sub-query $q^{(t)}$ from the original question and current memory. Second, \textbf{Dual-Stream Retrieval/Search} retrieves textual snippets $S^{(t)}$ and graph triples $R^{(t)}$. Third, \textbf{Response Generation} produces an intermediate answer or reasoning state $y^{(t)}$. Fourth, \textbf{Memory Synchronization} updates both memories through a bidirectional operator:
\begin{equation}
\bigl(\mathcal{T}^{(t+1)}, \mathcal{G}^{(t+1)}\bigr)
\leftarrow
\mathrm{Sync}\bigl(\mathcal{T}^{(t)}, \mathcal{G}^{(t)}, S^{(t)}, R^{(t)}, y^{(t)}\bigr)
\end{equation}
Co-E supports both MCTS and CoT reasoning modes; both use the same retrieval and synchronization machinery, differing only in how they choose the next sub-query.

\subsection{Shared Graph-Text Memory}

Co-E maintains a single shared memory object rather than independent per-hop contexts. Every step reads from the current memory and writes verified evidence back the same memory, so early evidence can affect later retrieval and generation.

\textbf{Textual memory.}
The textual memory $\mathcal{T}$ stores self-contained, question-relevant snippets: extracted evidence, intermediate sub-answers, verifier feedback, and natural-language statements produced from graph triples. Raw documents never enter memory; an extractor filters each retrieved document into atomic snippets.

\textbf{Graph memory.}
The graph memory $\mathcal{G}$ is a directed graph over entities and relations. \textit{KB-schema triples} are canonical and retrieved directly from a knowledge base. \textit{Open-vocabulary triples} are extracted from textual memory and use natural-language relation labels. The former provide reliable schema-grounded structure, while the latter let Co-E represent relations expressed in text but absent from the KB schema. A shared entity dictionary links surface mentions in $\mathcal{T}$ to KB identifiers in $\mathcal{G}$, allowing textual and graph evidence to refer to the same entities.

Before triples are merged into $\mathcal{G}$, a triple pruner removes irrelevant, redundant, or contradicted edges. Nodes that resolve to the same KB identifier are collapsed, and isolated low-value nodes are discarded. Thus graph memory is actively maintained rather than monotonically accumulated.

\subsection{Bidirectional Memory Synchronization}
\label{sec:sync}

The synchronization operator is the core mechanism of Co-E. It co-evolves text and graph memory through four operations.

\textbf{Textual consolidation.}
Co-E first consolidates $\mathcal{T}^{(t)}$, newly retrieved snippets $S^{(t)}$, and the intermediate response $y^{(t)}$. The consolidator deduplicates overlapping snippets, merges complementary evidence, and retracts contradicted intermediate predictions. This prevents the memory from becoming an ever-growing block of noisy context.

\textbf{Text-to-graph propagation.}
From the consolidated textual memory, an entity linker identifies entities and maps them to KB identifiers when possible. A relation extractor then proposes open-vocabulary triples from the text. These triples are combined with KB triples $R^{(t)}$, pruned for relevance and consistency, and merged into $\mathcal{G}^{(t)}$. This step lets newly discovered textual evidence expand the relational frontier.

\textbf{Graph-to-text propagation.}
Co-E textualizes surviving graph triples and appends them to textual memory as explicit relational statements. This gives the generator direct access to graph constraints in natural language, rather than requiring it to infer them from graph structure separately.

\textbf{Textual re-consolidation.}
Finally, the consolidator runs a second pass over the textual memory after graph-derived statements have been injected. The result is a coherent memory state $\mathcal{M}^{(t+1)}$ in which textual evidence and graph evidence are reconciled. This updated memory conditions the next sub-query, retrieval step, and response generation.

\subsection{Dual-Stream Retrieval}
\label{sec:retrieval}

Co-E retrieves from both text and graph sources: \textbf{Corpus retrieval.} 
A query generator rewrites $q^{(t)}$ into search-oriented queries conditioned on $\mathcal{M}^{(t)}$. These queries are sent to either web search or dense retrieval over a local corpus. Retrieved documents are passed to the extractor, which produces the snippet set $S^{(t)}$. Only atomic extracted snippets enter memory. \textbf{Graph retrieval.} Graph retrieval uses the current memory to choose which KB regions to explore. Co-E links entities from $q^{(t)}$ and $\mathcal{M}^{(t)}$ to KB identifiers, augments the seed set with relevant entities already present in $\mathcal{G}^{(t)}$, and issues $k$-hop SPARQL queries from this seed set. To control combinatorial growth, candidate triples are reranked between hops and then filtered by the triple pruner. The resulting triples form $R^{(t)}$. Because the seed set depends on the current graph memory, graph retrieval is memory-conditioned rather than a fixed precomputed traversal.

\subsection{Reasoning Modes}
\label{sec:modes}

Co-E can run the same memory update cycle under two reasoning policies.

\textbf{MCTS mode.}
MCTS mode searches over candidate reasoning trajectories. The root is the original question $q$, and child nodes are generated by the LLM. Following \cite{qi2024mutualreasoningmakessmaller}, nodes may represent sub-question answering, self-correction, synthesis, question rephrasing, or final-answer generation. Selection uses a prior-weighted UCB score, following the PUCT-style family \cite{Silver2017MasteringTG} rather than classic UCT:
\begin{equation}
S(s,a) =
\bar{Q}(s,a) + c \cdot P(s,a) \cdot
\frac{\sqrt{N_{\mathrm{p}}(s)+1}}{1 + N(s,a)}
\end{equation}
where $\bar{Q}(s,a)$ is the average reward accumulated by backpropagation, $P(s,a)$ is a child node type's heuristic prior, $N(s,a)$ is the child visit count, $N_{\mathrm{p}}(s)$ is the parent visit count, and $c$ controls exploration. Co-E separates tree statistics from memory state: node visits and values remain branch-specific, while evidence discovered during MCTS is accumulated in a shared working memory. After rollout, a verifier scores candidate terminal states under three views: no retrieved context, textual memory, and graph memory. Its natural-language assessments are added to textual memory as system-prediction items. After backpropagation, Co-E applies $\mathrm{Sync}$ once to the shared memory, where consolidation can remove unsupported predictions and filters newly introduced triples. This lets high-quality search feedback shape later retrieval, while relying on verifier feedback and active maintenance to reduce contamination from weak branches.

\textbf{CoT mode.}
CoT mode follows a sequential reasoning chain. At each step, the model generates and answers sub-questions conditioned on $q$ and $\mathcal{M}^{(t)}$, then applies the same synchronization operator before continuing. CoT mode is cheaper because it avoids tree search and verifier-guided branch selection, while MCTS mode explores multiple candidate reasoning paths before committing.

\section{Experiments}

\begin{table}[t]
\centering
\resizebox{\columnwidth}{!}{
\begin{tabular}{llccc}
\hline
Method & Backbone & Training & WebQSP & CWQ  \\
\hline
\multicolumn{5}{l}{\textbf{Learned/Subgraph KG-Retrieval KGQA}} \\
\textit{SubgraphRAG \cite{li2025simpleeffectiverolesgraphs}} & \textit{GPT-4o-mini/ GPT-4o} & \textit{MLP Retriever} & \textit{90.1} & \textit{66.7} \\
\textit{iQUEST \cite{wang-yu-2025-iquest}} & \textit{GPT-4o} & \textit{GNN} & \textit{88.9} & \textit{73.9} \\
GNN-RAG \cite{mavromatis-karypis-2025-gnn} & LLaMA-2-7B & GNN & \textbf{85.7} & 66.8 \\
\hline
\multicolumn{5}{l}{\textbf{Live-KG KGQA}} \\
\textit{ToG 2.0 \cite{ma2025thinkongraph20deepfaithful}} & \textit{GPT-3.5-turbo} & \textit{-} & \textit{81.1} & \textit{-} \\
\textit{PoG \cite{tan2025pathsovergraphknowledgegraphempowered}} & \textit{GPT-3.5-turbo} & \textit{-} & \textit{93.9} & \textit{74.7} \\
\textit{Interactive-KBQA \cite{xiong-etal-2024-interactive}} & \textit{GPT-4} & \textit{-} & \textit{72.5} & \textit{59.2} \\
ReKG-MCTS \cite{song-etal-2025-rekg} & LLaMA-3-8B & - & 72.2 & 59.8 \\
KBQA-o1 \cite{luo2025kbqao1agenticknowledgebase} & LLaMA-3.1-8B & - & 68.3 & 57.8 \\
MCTS-KBQA \cite{xiong2025mctskbqamontecarlotree} & LLaMA-3.1-8B & - & 73.5 & 72.1 \\
RoG \cite{luo2024reasoninggraphsfaithfulinterpretable} & LLaMA-2-7B & SFT & \textbf{85.7} & 62.6 \\
\hline
\multicolumn{5}{l}{\textbf{Ours}} \\
Co-E (CoT) & Qwen3-8B & - & 85.3 & 70.1 \\
Co-E (MCTS) & Qwen3-8B & - & 85.5 & \textbf{74.9} \\
\hline
\end{tabular}
}
\caption{KGQA Hits@1 on WebQSP and CWQ. \textit{Italic} rows use closed-source or $>$14B backbones; \textbf{bold} marks the best open $\le$8B score.}
\label{tab:result-kgqa}
\end{table}

\begin{table*}[t]
\centering
\resizebox{\textwidth}{!}{
\begin{tabular}{llccccc}
\hline
Method & Backbone & Training & HotpotQA & 2WikiMultiHopQA & MuSiQue & Bamboogle \\
\hline
\multicolumn{7}{l}{\textbf{Corpus-Based RAG}} \\
\textit{HopRAG \cite{liu2025hopragmultihopreasoninglogicaware}} & \textit{GPT-4o} & \textit{-} & \textit{62.0} & \textit{61.1} & \textit{42.2} & \textit{-} \\
\textit{Search-o1 \cite{li2025searcho1agenticsearchenhancedlarge}} & \textit{QwQ-32B} & \textit{-} & \textit{45.2} & \textit{58.0} & \textit{16.6} & \textit{56.0} \\
IRCoT \cite{trivedi-etal-2023-interleaving} & Qwen3-8B & - & 50.9 & 47.6 & 16.4 & 32.2 \\
Search-R1 \cite{jin2025searchr1trainingllmsreason} & Qwen2.5-7B & PPO & 43.3 & 38.2 & 19.6 & 43.2 \\
MR-Search \cite{xiao2026metareinforcementlearningselfreflectionagentic} & Qwen2.5-7B & Meta-RL & 46.8 & 43.6 & 22.1 & 45.2 \\
\hline
\multicolumn{7}{l}{\textbf{Graph-Augmented Retrieval}} \\
\textit{C2RAG \cite{ma2026mitigatingkgqualityissues}} & \textit{GPT-4o-mini} & \textit{-} & \textit{55.3} & \textit{65.9} & \textit{33.2} & \textit{-} \\
\textit{PAGER \cite{li2026structuredknowledgerepresentationcontextual}} & \textit{Qwen3-32B} & \textit{-} & \textit{50.6} & \textit{57.4} & \textit{23.0} & \textit{62.4} \\
GraphAnchor \cite{liu2026graphanchoredknowledgeindexingretrievalaugmented} & Qwen2.5-7B & - & 53.8 & 52.2 & 21.2 & 24.8 \\
Graph-R1 + EKA \cite{wang2026multihopreasoningearlyknowledge} & Qwen2.5-7B & GRPO & 59.4 & 60.9 & 40.6 & - \\
ProGraph-R1 \cite{park2026hypergraphproprogressawarereinforcementlearning} & Qwen2.5-7B & GRPO & 60.9 & 59.4 & 39.8 & - \\
\hline
\multicolumn{7}{l}{\textbf{Memory-Augmented}} \\
\textit{SubQRAG \cite{li2025subqragsubquestiondrivendynamic}} & \textit{GPT-4o-mini} & \textit{-} & \textit{56.0} & \textit{61.9} & \textit{29.7} & \textit{-} \\
SE-Search \cite{li2026sesearchselfevolvingsearchagent} & Qwen2.5-3B & GRPO & 45.0 & 36.1 & 18.3 & 42.4 \\
\hline
\multicolumn{7}{l}{\textbf{Ours}} \\
Co-E (CoT) & Qwen3-8B & - & 59.6 & 70.1 & 49.3 & 66.4 \\
Co-E (MCTS) & Qwen3-8B & - & \textbf{61.8} & \textbf{72.6} & \textbf{52.5} & \textbf{70.0} \\
\hline
\end{tabular}
}
\caption{Multi-hop text-QA EM. \textit{Italic} rows use closed-source or $>$14B backbones; \textbf{bold} marks the best open $\le$8B.}
\label{tab:result-textqa}
\end{table*}

\subsection{Experimental Setup}

\noindent \textbf{Datasets.}
We evaluate on six benchmarks spanning KGQA over structured knowledge bases and multi-hop text-QA over unstructured corpora. For KGQA, WebQSP \cite{yih2016value} contains mostly 1-2 hop Freebase questions, while CWQ \cite{talmor2018webknowledgebaseansweringcomplex} requires conjunction, composition, and longer relation chains. The main KGQA comparison follows the standard topic-subgraph protocol: each question is evaluated against a topic-centered Freebase candidate graph rather than against the full KB. For text-QA, 2WikiMultiHopQA \cite{ho-etal-2020-constructing} emphasizes entity-centric bridge reasoning, HotpotQA \cite{yang-etal-2018-hotpotqa} contains Wikipedia questions with stronger lexical cues, MuSiQue \cite{trivedi-etal-2022-musique} composes single-hop questions into longer chains, and Bamboogle \cite{press2023measuringnarrowingcompositionalitygap} contains adversarial multi-hop questions that stress bridge-entity reuse.

\noindent \textbf{Metrics.}
For KGQA, we report Hits@1, following prior work on WebQSP and CWQ \cite{ma2025thinkongraph20deepfaithful, li2025simpleeffectiverolesgraphs}; a prediction is correct if the top answer matches any gold alias after normalization. For text-QA, we report Exact Match (EM) \cite{trivedi-etal-2023-interleaving, liu2025hopragmultihopreasoninglogicaware}. Some reported baselines only release F1; these are labeled explicitly and treated as reference points rather than direct EM comparisons. Because EM penalizes correct paraphrases or formatting differences, ablations and Appendix~\ref{sec:full-results} also report LLM-judged accuracy (Acc) using Qwen3-30B-A3B-Thinking-2507 \cite{qwen3technicalreport}, which receives the question, gold answer, and prediction.

\noindent \textbf{Baselines.}
We group baselines by evidence representation. For text-QA, \textit{Corpus-Based RAG} methods rely primarily on textual retrieval and reasoning, including IRCoT \cite{trivedi-etal-2023-interleaving}, HopRAG \cite{liu2025hopragmultihopreasoninglogicaware}, ComposeRAG \cite{wu2025composeragmodularcomposablerag}, RT-RAG \cite{shi2026reasoningtreesimprovingretrievalaugmented}, Search-o1 \cite{li2025searcho1agenticsearchenhancedlarge}, Search-R1 \cite{jin2025searchr1trainingllmsreason}, MR-Search \cite{xiao2026metareinforcementlearningselfreflectionagentic}, and Search-P1 \cite{xia2026searchp1pathcentricrewardshaping}. \textit{Graph-Augmented Retrieval} includes corpus-graph and graph-trained systems such as HippoRAG 2 \cite{gutiérrez2025ragmemorynonparametriccontinual}, C2RAG \cite{ma2026mitigatingkgqualityissues}, PAGER \cite{li2026structuredknowledgerepresentationcontextual}, GraphAnchor \cite{liu2026graphanchoredknowledgeindexingretrievalaugmented}, EA-GraphRAG \cite{dong2026usegraphneedsefficiently}, Graph-R1 + EKA \cite{wang2026multihopreasoningearlyknowledge}, and ProGraph-R1 \cite{park2026hypergraphproprogressawarereinforcementlearning}. \textit{Memory-Augmented} systems include SubQRAG \cite{li2025subqragsubquestiondrivendynamic}, MemSearch-o1 \cite{zhang2026memsearcho1empoweringlargelanguage}, and SE-Search \cite{li2026sesearchselfevolvingsearchagent}. For KGQA, \textit{Learned/Subgraph KG-Retrieval KGQA} systems learn or compute query-specific candidate subgraphs before LLM reasoning, including SubgraphRAG \cite{li2025simpleeffectiverolesgraphs}, GNN-RAG \cite{mavromatis-karypis-2025-gnn}, and iQUEST \cite{wang-yu-2025-iquest}. \textit{Live-KG KGQA} systems traverse or query a knowledge graph at inference time, including ToG 2.0 \cite{ma2025thinkongraph20deepfaithful}, PoG \cite{tan2025pathsovergraphknowledgegraphempowered}, Interactive-KBQA \cite{xiong-etal-2024-interactive}, MCTS-KBQA \cite{xiong2025mctskbqamontecarlotree}, ReKG-MCTS \cite{song-etal-2025-rekg}, KBQA-o1 \cite{luo2025kbqao1agenticknowledgebase}, RoG \cite{luo2024reasoninggraphsfaithfulinterpretable}, and KERAG \cite{sun2025kerag}. Open $\le$8B models are our primary comparison group; closed-source and much larger models are reference.

\noindent \textbf{Implementation Details.}
Co-E is training-free. We use Qwen3-8B \cite{qwen3technicalreport} as the reasoning backbone, Qwen3-Embedding-4B \cite{qwen3embedding} for dense retrieval over the Wiki23 corpus, and SPARQL for graph retrieval. The same graph-text memory mechanism is used across settings, but the graph source differs. In the main KGQA experiments, graph retrieval is restricted to the benchmark-provided Freebase candidate subgraph. In text-QA, graph retrieval uses full live-Wikidata. For detail implementation, see Appendix~\ref{sec:appendix_implementation}.

\subsection{KGQA Results}

Table~\ref{tab:result-kgqa} reports Hits@1 on WebQSP and CWQ under the standard topic-subgraph protocol. Each question is paired with a compact Freebase subgraph centered on the topic entity. This setting tests reasoning over a noisy but bounded candidate graph: Co-E does not expand the Freebase search space beyond the provided subgraph, but it can prune candidate relations, preserve intermediate entities, and textualize graph facts for later generation. On WebQSP, where questions are short 1-2 hop queries and the answer path is often exposed by the topic graph, the highest scores come from closed-source or trained systems such as PoG, SubgraphRAG, and iQUEST. Co-E (MCTS) reaches 85.5 Hits@1, below those systems but essentially tied with the strongest open 7B trained baselines, RoG and GNN-RAG (both 85.7). Co-E (CoT) reaches a similar 85.3, suggesting that synchronized memory, rather than tree search alone, accounts for most of Co-E's WebQSP performance. On CWQ, the advantage of maintaining graph-text state is clearer because questions require conjunctions, composition, and longer relation chains. Co-E (MCTS) reaches 74.9 Hits@1, slightly above PoG (74.7) and iQUEST (73.9), and improves over MCTS-KBQA by 2.8 points, GNN-RAG by 8.1 points, SubgraphRAG by 8.2 points, RoG by 12.3 points, and ReKG-MCTS by 15.1 points. The larger MCTS-CoT gap on CWQ indicates that search helps when several plausible relation paths compete. Overall, the KGQA results support Co-E as a training-free graph-memory reasoner, with the largest gains when the model must preserve and select among multi-step relations.

\begin{table}[t]
\centering
\resizebox{\columnwidth}{!}{
\begin{tabular}{lcccc}
\hline
\multirow{2}{*}{Variant} & \multicolumn{2}{c}{MCTS} & \multicolumn{2}{c}{CoT} \\
\cmidrule(lr){2-3} \cmidrule(lr){4-5}
 & EM & Acc & EM & Acc \\
\hline
Co-E (full) & \textbf{70.0} & \textbf{77.6} & \textbf{66.4} & \textbf{75.2} \\
\hline
\multicolumn{5}{l}{\textbf{Memory synchronization}} \\
$-$ Bidirectional sync & 61.1 & 65.9 & 52.5 & 55.7 \\
$-$ Unidirectional (text$\rightarrow$graph) & 64.0 & 71.5 & 56.1 & 61.6 \\
$-$ Unidirectional (graph$\rightarrow$text) & 63.1 & 70.3 & 55.9 & 60.9 \\
$-$ Active maintenance & 49.5 & 59.7 & 49.0 & 47.1 \\
\hline
\multicolumn{5}{l}{\textbf{Memory modality}} \\
$-$ Graph memory (text only) & 54.8 & 62.3 & 50.3 & 54.9 \\
$-$ Textual memory (graph only) & 55.0 & 63.6 & 50.6 & 54.7 \\
\hline
\multicolumn{5}{l}{\textbf{Retrieval and extraction}} \\
$-$ Extractor & 50.2 & 61.3 & 48.1 & 50.9 \\
$-$ Graph retriever & 49.8 & 54.2 & 49.9 & 50.8 \\
$-$ Textual retriever & 46.6 & 50.3 & 45.8 & 50.1 \\
\hline
\end{tabular}
}
\caption{Bamboogle ablations for MCTS and CoT.}
\label{tab:ablation}
\end{table}

\subsection{Multi-hop Text-QA Results}

Table~\ref{tab:result-textqa} reports EM on the four text-QA benchmarks. Co-E (MCTS) is strongest among open $\le$8B systems, reaching 61.8 on HotpotQA, 72.6 on 2WikiMultiHopQA, 52.5 on MuSiQue, and 70.0 on Bamboogle. Co-E (CoT) is lower than MCTS but remains competitive with same scale baselines, suggesting that synchronized memory provides much of the improvement while tree search adds gain when bridge candidates compete.

The largest gains appear on datasets where a later hop depends on preserving a bridge entity or relation discovered earlier. On 2WikiMultiHopQA, Co-E improves over C2RAG by 6.7 points and SubQRAG by 10.7 points. This supports the intended use of text-to-graph propagation: once a passage identifies a bridge entity, Co-E can link it, store it as graph memory, and use it to seed later graph or corpus retrieval. The advantage is clearer on harder compositional text-QA benchmarks. On MuSiQue, Co-E reaches 52.5 EM, outperforming HopRAG-GPT-4o by 10.3 points, ProGraph-R1 by 12.7 points, and SubQRAG by 22.8 points. MuSiQue's longer 2-4 hop chains make retrieval noise compound; active consolidation and graph-text synchronization preserve the resolved chain while discarding stale content. Bamboogle shows a similar pattern: Co-E reaches 70.0 EM, improving over Search-R1-7B by 26.8 points and PAGER-32B by 7.6 points. These questions often cannot be answered until an intermediate entity is resolved, so MCTS explores alternative bridge candidates while synchronization commits verified evidence into both memories. HotpotQA has the smallest margin: Co-E reaches 61.8 EM, ahead of SubQRAG and ProGraph-R1 but close to HopRAG-GPT-4o. HotpotQA often has stronger lexical cues and shorter chains, making corpus-only or closed-source pipelines more competitive.

\begin{table}[t]
\centering
\resizebox{\columnwidth}{!}{
\begin{tabular}{lcc}
\hline
Method & WebQSP Hits@1 & CWQ Hits@1 \\
\hline
Co-E (subgraph protocol) & 85.5 & 74.9 \\
Co-E (Full-Wikidata) & 78.6 & 73.4 \\
\hline
\end{tabular}
}
\caption{Co-E with Freebase topic-subgraph versus Full-Wikidata retrieval (Hits@1).}
\label{tab:full_wikidata}
\end{table}

\begin{figure}[t]
  \centering
  \includegraphics[width=0.45\textwidth]{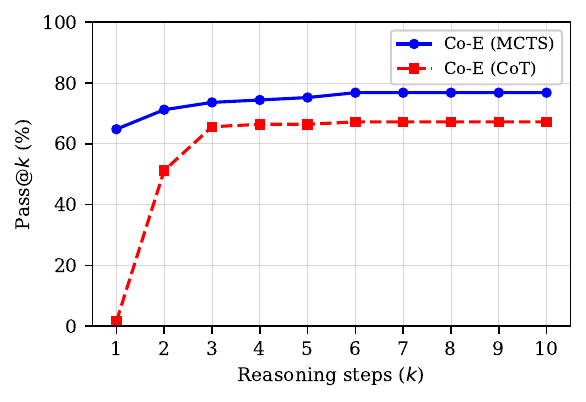}
  \caption{Pass@$k$ on Bamboogle; $k$ is the number of tree expansions for MCTS and chain depth for CoT.}
  \label{fig:pass_at_k}
\end{figure}

\subsection{Ablation Study}

Table~\ref{tab:ablation} ablates Co-E on Bamboogle, where questions often require a resolved bridge entity in a later hop. The ablations test three design choices: whether synchronization must be bidirectional, whether both memory modalities are needed, and whether retrieval-time filtering can replace persistent memory.

\textbf{Synchronization.} Removing bidirectional synchronization drops MCTS EM from 70.0 to 61.1 and CoT EM from 66.4 to 52.5. The drop under both search policies shows that the effect is not merely a byproduct of tree exploration. One-way propagation is also insufficient: text$\rightarrow$graph reaches 64.0 MCTS EM, while graph$\rightarrow$text reaches 63.1. These variants preserve one half of the loop, but lose the feedback cycle that lets textual evidence expand the graph frontier and graph facts shape the next textual context. Removing active maintenance causes the largest synchronization-related degradation, reducing MCTS EM to 49.5. This variant can still propagate evidence across modalities, but it no longer reliably deduplicates, merges, or retracts stale content. The result suggests that co-evolution must be selective rather than purely accumulative.

\textbf{Memory modality.} Removing graph memory lowers MCTS EM to 54.8, and removing textual memory lowers it to 55.0. The degradation suggests that textual memory provides grounding that KB triples may not encode, while graph memory preserves explicit entity-relation structure that text leaves implicit. Neither memory alone recovers the full behavior, supporting paired graph-text state.

\textbf{Retrieval and extraction.} Removing textual retrieval causes the largest component drop ($-$23.4 MCTS EM), followed by graph retrieval ($-$20.2) and the extractor ($-$19.8). This confirms that Co-E depends on both evidence streams, but also that retrieval alone is not enough. The extractor ablation shows that raw documents are too noisy to accumulate directly; filtering them into concise, self-contained snippets is part of memory construction rather than a minor preprocessing detail.

\begin{table*}[t]
\centering
\begin{minipage}[t]{0.52\textwidth}
  \centering
  \resizebox{\textwidth}{!}{
  \begin{tabular}{lp{4cm}p{9cm}}
  \toprule
  \textbf{Step} & \textbf{Sub-query} & \textbf{Textual memory after sync} \\
  \midrule
  \textbf{1} & Who is George Washington's father? & [Retrieval] George Washington's father was Augustine Washington. \\
  \addlinespace
  \textbf{2} & Who is Augustine Washington's mother? \textit{(graph-guided)} & [Retrieval] Augustine Washington was born to Mildred Warner and Capt. Lawrence Washington. \par [Graph] Augustine Washington -\textbf{mother}$\rightarrow$ Mildred Warner \\
  \bottomrule
  \end{tabular}
  }
\end{minipage}
\hfill
\begin{minipage}[c]{0.46\textwidth}
  \centering
  \includegraphics[width=\textwidth]{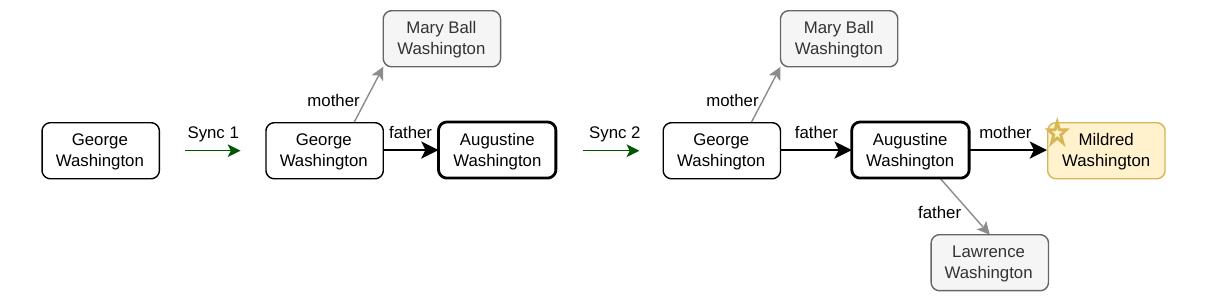}
\end{minipage}
\caption{Memory walkthrough for \textit{"Who is the mother of the father of George Washington?"}. Graph-injected text grounds step~2 retrieval on Augustine Washington; dead-end graph branches are gray and $\star$ marks the gold answer.}
\label{tab:memory_walkthrough}
\end{table*}

\subsection{Full-Wikidata Evaluation}
\label{sec:full-wikidata}

The KGQA results above follow the standard topic-subgraph protocol, where each question is paired with a pre-retrieved Freebase subgraph centered on its topic entity. This protocol enables controlled comparison with prior KGQA systems, but it assumes question-specific preprocessing: a topic entity must be identified and a compact candidate graph extracted before reasoning begins. Such preprocessing is not always available in open-domain settings, where systems may need to start from a large KG and from entities discovered during retrieval. To test this harder setting, we rerun Co-E (MCTS) on WebQSP and CWQ using the same Full-Wikidata retrieval stack as the text-QA experiments. This experiment is a stress test rather than a matched replacement for Freebase evaluation, since entity identifiers, relation schemas, and answer aliases differ between Freebase and Wikidata. Table~\ref{tab:full_wikidata} shows that moving from Freebase topic subgraphs to full Wikidata has only a small effect on CWQ, reducing Hits@1 from 74.9 to 73.4. For compositional questions, memory-conditioned entity discovery and SPARQL expansion can recover much of the useful relational frontier even without a benchmark-provided candidate graph. The drop is larger on WebQSP, from 85.5 to 78.6, because many WebQSP questions are short and the preprocessed subgraph often exposes the answer path directly (due to its significant smaller size and better schema match). Without this pruning, retrieval noise has a larger effect on short questions with less compositional structure. Overall, the Full-Wikidata setting shows that Co-E benefits from topic-subgraph pruning when available, while retaining strong compositional KGQA performance when forced to retrieve from a much larger and schema-mismatched graph.

\subsection{Analysis}

\textbf{Pass@$k$.} Figure~\ref{fig:pass_at_k} separates search from memory. MCTS improves rapidly in the first few expansions because it can evaluate multiple bridge candidates before committing, whereas CoT must extend a single chain. The CoT curve also improves as memory accumulates, indicating that synchronization strengthens later reasoning even without branch exploration. Both curves eventually plateau, suggesting that most gains come from early memory updates and that the remaining MCTS advantage reflects exploration over alternative bridge entities or relation chains.

\noindent \textbf{Memory evolution.} Table~\ref{tab:memory_walkthrough} illustrates how bidirectional synchronization changes the next retrieval action, not just the final context. For the question \textit{"Who is the mother of the father of George Washington?"}, the first step retrieves Augustine Washington as George Washington's father. Text$\rightarrow$graph propagation turns this evidence into parentage edges, and graph$\rightarrow$text propagation writes the resolved father relation back into textual memory. The second step can therefore query around Augustine Washington as a grounded entity rather than around an underspecified phrase. This steers SPARQL expansion toward Augustine Washington's parentage triples and injects the mother edge to Mildred Warner back into memory. The example also clarifies the role of pruning and search: memory may contain plausible off-path facts, while verification and branch selection decide which path answers the question.

\noindent \textbf{Failure analysis.} Manual inspection of Bamboogle failures, detailed in Appendix~\ref{sec:appendix_failures}, shows that remaining errors usually come from evidence quality or path selection rather than from the absence of graph-text memory. Retrieval gaps account for 45\% of failures: the system often identifies the right entity chain, but the corpus repeatedly supports an incorrect value, which synchronization cannot fix when the evidence is internally consistent. Incomplete reasoning chains account for 21\%, usually when relevant facts are present in memory but the synthesizer selects a chain that is too short, too long, or off by one hop. Wrong entity linking and synthesis errors each account for 14\%, and surface-form mismatch accounts for 7\%. These categories suggest remedies: stronger source cross-checking for retrieval gaps, better entity disambiguation for linking errors, and explicit chain-completeness verification for incomplete reasoning.

\section{Conclusion}

We introduced Co-E, a training-free multi-hop QA system built around synchronized graph-text working memory. Co-E treats reasoning as an iterative memory-maintenance problem: textual evidence can expand the graph frontier when open graph retrieval is available, and graph facts can be written back into the textual context that conditions later retrieval and generation. This bidirectional synchronization lets the system revise its relational state during inference rather than relying on a static retrieved context or a fixed topic-centered candidate graph. Across six KGQA and text-QA benchmarks, Co-E consistently improves over comparable training-free open-backbone baselines, achieving 72.6 EM on 2WikiMultiHopQA, 70.0 EM on Bamboogle, 52.5 EM on MuSiQue, and 74.9 Hits@1 on CWQ. Ablations show that both propagation directions and active memory maintenance are necessary: simply accumulating evidence is not enough. The main KGQA results show strong performance under the standard Freebase topic-subgraph protocol, while the Full-Wikidata stress test shows that Co-E can retain strong compositional KGQA performance without benchmark-specific subgraph preprocessing. Together, these results support graph-text co-evolution as a practical mechanism for training-free multi-hop reasoning.

\section*{Limitations}

Co-E is limited by the quality of the evidence it retrieves. Bidirectional synchronization can reconcile redundant or contradicted memory entries, but it cannot reliably correct a wrong fact that is repeatedly supported by the retrieved corpus. This appears in the Bamboogle failure analysis, where retrieval gaps account for 45\% of errors. Future systems could cross-check numerical attributes, superlatives, and entity facts against independent sources, though doing so would add latency. Co-E also depends on accurate entity linking and path selection. Wrong entity linking accounts for 14\% of Bamboogle failures, and incomplete reasoning chains account for 21\%. These errors suggest that graph-text memory is not sufficient by itself: the system still needs better candidate disambiguation, stronger hop-count constraints, and more explicit verification of whether the selected chain fully answers the question. MCTS mode uses a shared global working memory rather than a fully branch-local memory. Verifier assessments, consolidation, and graph pruning mitigate contamination from weak branches, but they do not provide rollback semantics; future work could compare shared-memory MCTS with branch-local or reward-gated memory commits. Finally, Co-E trades additional inference cost for stronger reasoning. MCTS mode requires multiple LLM calls per expansion; CoT mode is cheaper but less accurate. Stronger early stopping, cached verification, or adaptive switching between CoT and MCTS could reduce this cost. Our experiments primarily use a Qwen3-8B backbone, with a Bamboogle-only Qwen3.5-4B check in Appendix~\ref{sec:appendix_cost}; evaluating larger or more specialized models, and extending the smaller-backbone study across all benchmarks, remains future work.


\bibliography{custom}

\appendix

\section{Full Baseline Results}
\label{sec:full-results}
Tables~\ref{tab:appendix_kgqa} and \ref{tab:appendix_textqa} provide the full baseline comparison used to construct the main result tables. We include each method's backbone, training setting, reported metric, and any supplementary semantic-accuracy score when available. The tables include both training-free and trained systems, with closed-source or substantially larger backbones retained as reference points rather than primary comparisons.

\begin{table*}[h]
\centering
\resizebox{\textwidth}{!}{
\begin{tabular}{llcccl}
\toprule
Method & Backbone & Training & WebQSP & CWQ & Metric note \\
\midrule
\multicolumn{6}{l}{\textbf{Learned/Subgraph KG-Retrieval KGQA}} \\
SubgraphRAG \cite{li2025simpleeffectiverolesgraphs} & GPT-4o-mini/ GPT-4o & Yes (MLP) & 90.1 / 77.5 & 66.7 / 59.1 & Hits@1 / F1 \\
GNN-RAG \cite{mavromatis-karypis-2025-gnn} & LLaMA-2-7B & Yes (GNN) & 85.7 / 71.3 & 66.8 / 60.4 & Hits@1 / F1 \\
iQUEST \cite{wang-yu-2025-iquest} & GPT-4o & Yes (GNN) & 88.93 & 73.85 & Hits@1 \\
\midrule
\multicolumn{6}{l}{\textbf{Live-KG KGQA}} \\
ToG \cite{sun2024thinkongraphdeepresponsiblereasoning} & GPT-3.5-turbo & No & 76.2 & 57.1 & Hits@1 \\
ToG 2.0 \cite{ma2025thinkongraph20deepfaithful} & GPT-3.5-turbo & No & 81.1 & — & Hits@1 \\
PoG \cite{tan2025pathsovergraphknowledgegraphempowered} & GPT-3.5-turbo & No & 93.9 & 74.7 & Hits@1 \\
PPoGA \cite{jeon2025ppogapredictiveplanongraphaction} & GPT-3.5-turbo & No & 83.1 & 64.5 & Hits@1 \\
Interactive-KBQA \cite{xiong-etal-2024-interactive} & GPT-4 & No & 72.47 / 71.20 & 59.17 / 49.07 & Hits@1 / F1 \\
ReKG-MCTS \cite{song-etal-2025-rekg} & LLaMA-3-8B & No & 72.20 & 59.80 & Hits@1 \\
KBQA-o1 \cite{luo2025kbqao1agenticknowledgebase} & LLaMA-3.1-8B & No & 68.3$^*$ / 59.8 & 57.8$^*$ / 43.1$^*$ & Hits@1 / F1 \\
MCTS-KBQA \cite{xiong2025mctskbqamontecarlotree} & LLaMA-3.1-8B & No & 73.5 / 72.5 & 72.1 / 64.9 & Hits@1 / F1 \\
KERAG \cite{sun2025kerag} & LLaMA-3.1-70B & Yes (SFT)$^+$ & 84.3 & 70.20 & Hits@1 \\
RoG \cite{luo2024reasoninggraphsfaithfulinterpretable} & LLaMA-2-7B & Yes (SFT) & 85.7 / 70.8 & 62.6 / 56.2 & Hits@1 / F1 \\
KBQA-R1 \cite{sun2026kbqar1reinforcinglargelanguage} & LLaMA-3.1-8B & Yes (GRPO) & 83.4 & — & F1 \\
\midrule
Co-E (MCTS) & Qwen3-8B & No & 85.5 / 94.55 & 74.9 / 89.85 & Hits@1 / Acc \\
Co-E (CoT) & Qwen3-8B & No & 85.3 / 92.7 & 70.1 / 87.78 & Hits@1 / Acc \\
\bottomrule
\end{tabular}
}
\caption{Full KGQA baseline results. $^*$ denotes reproduced results. $^+$ KERAG trains an 8B model to summarize retrieved evidence.}
\label{tab:appendix_kgqa}
\end{table*}

\begin{table*}[h]
\centering
\resizebox{\textwidth}{!}{
\begin{tabular}{llcccccc}
\toprule
Method & Backbone & Training & HotpotQA & 2WikiMultiHopQA & MuSiQue & Bamboogle & Metric note \\
\midrule
\multicolumn{8}{l}{\textbf{Corpus-Based RAG}} \\
HopRAG (GPT-3.5) \cite{liu2025hopragmultihopreasoninglogicaware} & GPT-3.5-turbo & No & 61.3 / 78.3 & 61.6 / 68.9 & 39.1 / 53 & — & EM / F1 \\
HopRAG (GPT-4o) \cite{liu2025hopragmultihopreasoninglogicaware} & GPT-4o & No & 62 / 76.1 & 61.1 / 68.3 & 42.2 / 54.9 & — & EM / F1 \\
ComposeRAG \cite{wu2025composeragmodularcomposablerag} & GPT-4o & No & 58 / 73.2 & 80.8 / 82 & 34.2 / 38.8 & 62.4 / 72.8 & EM / Acc \\
RT-RAG (GPT-4o-mini) \cite{shi2026reasoningtreesimprovingretrievalaugmented} & GPT-4o-mini & No & 52.50 / 65.26 & 63.00 / 75.08 & 41.50 / 54.42 & — & EM / F1 \\
RT-RAG (Qwen2.5-14B) \cite{shi2026reasoningtreesimprovingretrievalaugmented} & Qwen2.5-14B & No & 51.00 / 66.24 & 64.00 / 73.69 & 39.00 / 50.04 & — & EM / F1 \\
Search-o1 \cite{li2025searcho1agenticsearchenhancedlarge} & QwQ-32B & No & 45.2 / 57.3 & 58.0 / 71.4 & 16.6 / 28.2 & 56.0 / 67.8 & EM / F1 \\
IRCoT \cite{trivedi-etal-2023-interleaving} & Qwen3-8B & No & 50.9 / 51.8 & 47.6 / 48.3 & 16.4 / 17.8 & 32.2 / 33.8 & EM / Acc \\
Search-R1 (7B) \cite{jin2025searchr1trainingllmsreason} & Qwen2.5-7B & Yes (PPO) & 43.30 & 38.20 & 19.6 & 43.2 & EM \\
Search-R1 (14B) \cite{jin2025searchr1trainingllmsreason} & Qwen2.5-14B & Yes (PPO) & 46.8 & 47 & 24.1 & 52.8 & EM \\
Search-P1 (7B) \cite{xia2026searchp1pathcentricrewardshaping} & Qwen2.5-7B & Yes (Dual-RL) & 42.9 & 39.8 & 21.8 & 44 & Acc \\
MR-Search (7B) \cite{xiao2026metareinforcementlearningselfreflectionagentic} & Qwen2.5-7B & Yes (Meta-RL) & 46.8 & 43.6 & 22.1 & 45.2 & EM \\
\midrule
\multicolumn{8}{l}{\textbf{Graph-Augmented Retrieval}} \\
HippoRAG 2 \cite{gutiérrez2025ragmemorynonparametriccontinual} & GPT-4o-mini & No & 56.3 / 71.1 & 60.5 / 69.7 & 35.0 / 49.3 & — & EM / F1 \\
C2RAG \cite{ma2026mitigatingkgqualityissues} & GPT-4o-mini & No & 55.3 / 72.8 & 65.9 / 73.9 & 33.2 / 46.1 & — & EM / F1 \\
PAGER \cite{li2026structuredknowledgerepresentationcontextual} & Qwen3-32B & No & 50.60 & 57.40 & 23.0 & 62.4 & EM \\
GraphAnchor \cite{liu2026graphanchoredknowledgeindexingretrievalaugmented} & Qwen2.5-7B & No & 53.8 / 66.0 & 52.2 / 61.6 & 21.2 / 32.6 & 24.8 / 34.3 & EM / F1 \\
EA-GraphRAG \cite{dong2026usegraphneedsefficiently} & GPT-4o-mini & No & 65.9 / 80.2 & 76.3 / 81.5 & — & — & EM / Acc \\
Graph-R1 + EKA \cite{wang2026multihopreasoningearlyknowledge} & Qwen2.5-7B & Yes (GRPO) & 59.38 / 66.14 & 60.94 / 68.26 & 40.63 / 51.63 & — & EM / F1 \\
ProGraph-R1 (3B) \cite{park2026hypergraphproprogressawarereinforcementlearning} &  Qwen2.5-3B & Yes (GRPO) & 55.5 / 61.9 & 55.5 / 61.7 & 37.5 / 47.3 & — & EM / F1 \\
ProGraph-R1 (7B) \cite{park2026hypergraphproprogressawarereinforcementlearning} & Qwen2.5-7B & Yes (GRPO) & 60.9 / 67.6 & 59.4 / 69.8 & 39.8 / 49.5 & — & EM / F1 \\
\midrule
\multicolumn{8}{l}{\textbf{Memory-Augmented}} \\
SubQRAG \cite{li2025subqragsubquestiondrivendynamic} & GPT-4o-mini & No & 56.00 / 64.30 & 61.90 / 64.30 & 29.70 / 38.14 & — & EM / F1 \\
MemSearch-o1 \cite{zhang2026memsearcho1empoweringlargelanguage} & Qwen2.5-72B-Instruct & No & 59.71 & 65.95 & 44.11 & — & F1 \\
MemSearch-o1 \cite{zhang2026memsearcho1empoweringlargelanguage} & DeepSeek-V3.1 & No & 67.78 & 68.32 & 52.01 & — & F1 \\
SE-Search \cite{li2026sesearchselfevolvingsearchagent} & Qwen2.5-3B & Yes (GRPO) & 45.0 & 36.10 & 18.3 & 42.4 & EM \\
\midrule
Co-E (MCTS) & Qwen3-8B & No & 61.8 / 86.17 & 72.6 / 81.6 & 52.5 / 60.9 & 70.0 / 77.6 & EM / Acc \\
Co-E (CoT) & Qwen3-8B & No & 59.6 / 83.48 & 70.1 / 80.6 & 49.3 / 60.3 & 66.4 / 75.2 & EM / Acc \\
\bottomrule
\end{tabular}
}
\caption{Full multi-hop text-QA baseline results.}
\label{tab:appendix_textqa}
\end{table*}

\section{Implementation Details}
\label{sec:appendix_implementation}
LLM inference runs on 4 NVIDIA A100 80GB GPUs. The Qwen3-8B reasoning backbone is served on 2 GPUs with SGLang model parallelism and thinking enabled, then exposed through a LiteLLM OpenAI-compatible proxy. Embedding and reranking models run on a separate GPU, while evaluation runs on a CPU server with 128 GB RAM and 32 cores. Dense retrieval uses an in-memory FAISS index over the Wiki23 corpus. SPARQL queries target either a local Virtuoso endpoint for KGQA or the public Wikidata endpoint for text-QA. Redis caches repeated calls during MCTS simulations.

\begin{table}[h]
\centering
\resizebox{0.45\textwidth}{!}{%
\begin{tabular}{lll}
\toprule
Group & Parameter & Value \\
\midrule
\textbf{LLM Generation} & Backbone & Qwen3-8B \\
 & Temperature & 1.0 \\
 & top-p & 0.95 \\
 & top-k & 20 \\
 & Presence penalty & 1.5 \\
 & Thinking & enabled \\
\textbf{Retrieval} & Embedding model & Qwen3-Embedding-4B \\
 & Reranker model & Qwen3-Reranker-4B \\
 & Reranker top-k & 10 \\
 & Web search top-k & 5 \\
 & Graph hops ($k$) & 2 \\
 & Entity candidates per mention & 3 \\
 & Triple pruner top-k & 128 \\
\textbf{MCTS} & Max iterations & 20 \\
 & Max tree depth & 10 \\
 & Exploration weight ($c$) & 2.5 \\
 & Max simulation depth & 5 \\
 & Min iterations before early stop & 5 \\
 & High-confidence threshold & 0.9 \\
 & Convergence patience & 5 \\
 & Semantic sufficiency count & 5 \\
\textbf{MCTS node priors} & Sub-question/Answer & 0.60 \\
 & Self-Correction & 0.50 \\
 & Synthesis & 0.45 \\
 & Question Rephrasing & 0.40 \\
 & Final Answer & 0.30 \\
\textbf{CoT} & Max chain depth & 10 \\
 & Max sub-questions per step & 3 \\
\textbf{Working Memory} & Max textual memory tokens & 8192 \\
\bottomrule
\end{tabular}%
}
\caption{Configuration hyperparameters applied across the Co-E framework.}
\end{table}

\subsection{Knowledge Base Setup}

\textbf{KGQA benchmarks (WebQSP, CWQ).} We follow the standard topic-subgraph protocol: each question is paired with a pre-retrieved Freebase candidate subgraph centered on its topic entity, and graph retrieval is restricted to that subgraph. This reduces the full-KG search space but still leaves noisy and off-path relations for the system to filter. SPARQL queries target a local Virtuoso endpoint loaded with the per-question subgraphs. Entity linking maps surface mentions to Freebase MIDs using a dictionary derived from the benchmark training set; no new entities are added during evaluation. Corpus retrieval uses FAISS over the Wiki23 corpus with Qwen3-Embedding-4B.

\textbf{Text-QA benchmarks (2WikiMultiHopQA, HotpotQA, MuSiQue, Bamboogle).} We use full Wikidata without pre-selection. Graph retrieval issues $k$-hop SPARQL queries against the public Wikidata endpoint and prunes candidates to the top 128 triples. Entity linking maps surface mentions to Wikidata QIDs; unlike the KGQA setting, the shared entity dictionary grows as new entities are discovered. Corpus retrieval again uses FAISS over the Wiki23 corpus.

\textbf{Full-Wikidata KGQA.} The Full-Wikidata evaluation in \S~\ref{sec:full-wikidata} applies the text-QA protocol to WebQSP and CWQ: the same Wikidata endpoint, no pre-selected subgraph, and the same dual-stream retrieval stack. This setting tests robustness when the knowledge-base source is not benchmark-adapted.

\section{Inference Cost Analysis}
\label{sec:appendix_cost}

Table~\ref{tab:cost_per_role} reports role-level inference cost on Bamboogle: average LLM calls per question, output (non-thinking) tokens per call, and total output tokens per question. These counts include every LLM invocation, both cache hits and misses; our implementation caches repeated LLM and retrieval calls, which reduces wall-clock latency but does not lower the totals reported here, since they reflect the underlying compute demand. Co-E (MCTS) uses 204.1 calls and 27.0K output tokens per question, compared with 58.0 calls and 7.8K tokens for Co-E (CoT), and 9.3 calls and 1.2K tokens for IR-CoT. In output tokens, MCTS is therefore about 3.5$\times$ CoT and 22$\times$ IR-CoT. The cost is concentrated in retrieval-side construction, not reasoning. Retrieval roles account for 57.1\% of MCTS output tokens and 68.7\% of CoT output tokens. Within retrieval, the extractor is the largest contributor at 6.2K tokens per MCTS question and 2.7K per CoT question; within reasoning, answer generation contributes 8.0K and 2.1K tokens. MCTS further repeats this loop over alternative branches and adds verifier calls. The added cost is therefore tied to making intermediate evidence explicit, pruning noisy state, and exploring multiple reasoning paths.

To probe how much of Co-E's accuracy depends on the 8B backbone, we re-run both modes on Bamboogle using Qwen3.5-4B \cite{qwen3.5} as the backbone for every role. Results are shown in Table~\ref{tab:backbone_size}. The smaller backbone loses 1.8 EM in MCTS mode and 3.3 EM in CoT mode. Co-E therefore degrades gracefully when the backbone is roughly halved, consistent with the design hypothesis that most of the gain comes from synchronized graph-text memory rather than from raw backbone capacity. This also provides direct evidence for the deployment argument below: in settings where 8B inference is too expensive, a 4B-class backbone preserves most of the accuracy at a fraction of the per-call compute.

\begin{table}[h]
\centering
\small
\begin{tabular}{lcc}
\toprule
Backbone & MCTS (EM / Acc) & CoT (EM / Acc) \\
\midrule
Qwen3-8B    & 70.0 / 77.6 & 66.4 / 75.2 \\
Qwen3.5-4B  & 68.2 / 74.9 & 63.1 / 69.5 \\
$\Delta$    & $-1.8$ / $-2.7$ & $-3.3$ / $-5.7$ \\
\bottomrule
\end{tabular}
\caption{Bamboogle backbone-size sensitivity for Co-E.}
\label{tab:backbone_size}
\end{table}

This cost can be reduced without changing the training-free setup. We use Qwen3-8B for all roles because it is a relatively small, locally deployable backbone and keeps evaluation controlled. The Qwen3.5-4B result above shows that a roughly halved backbone preserves most of Co-E's Bamboogle accuracy, so the cheaper-backbone direction is empirically supported and not just hypothetical. In deployment, Co-E's role-based prompting makes model routing straightforward: retrieval-side roles such as NER, extraction, and pruning can use cheaper 2B-4B models or deterministic tools, while stronger models are reserved for harder reasoning roles such as question generation, verification, and consolidation. Thus, inference cost can be significantly reduced because most computation is on the retrieval side. Caching can also reuse repeated entity-linking, SPARQL, extraction, verification, and LLM-call outputs. Finally, CoT provides the same graph-text memory mechanism at much lower cost, while MCTS can be reserved for ambiguous or long-chain questions.

\begin{table*}[h]
\centering
\scriptsize
\resizebox{\textwidth}{!}{
\begin{tabular}{llcccc}
\toprule
Role & Category & MCTS calls/q & CoT calls/q & IR-CoT calls/q &  Toks/call \\
\midrule
ANSWER\_GENERATOR & Reasoning & 33.3 & 8.9 & -- & 240.4 \\
VERIFIER & Reasoning & 14.8 & -- & -- & 79.4 \\
SUBQUESTION\_GENERATOR & Reasoning & 8.4 & 1.4 & -- & 203.7 \\
SELF\_CORRECTOR & Reasoning & 3.7 & -- & -- & 118.3 \\
FINAL\_ANSWER\_SYNTHESIZER & Reasoning & 1.0 & -- & -- & 280.5 \\
EXTRACTOR & Retrieval & 43.8 & 19.4 & -- & 140.6 \\
TRIPLE\_PRUNER (KB retrieval) & Retrieval & 17.5 & 5.8 & -- & 30.7 \\
QUERY\_GENERATOR & Retrieval & 28.8 & 7.9 & -- & 129.6 \\
NER (entity linking) & Retrieval & 28.8 & 7.9 & -- & 61.1 \\
MEMORY\_CONSOLIDATOR & Retrieval & 8.7 & 2.5 & -- & 221.8 \\
NER (memory sync) & Retrieval & 5.1 & 1.4 & -- & 56.2 \\
RELATION\_EXTRACTOR & Retrieval & 5.1 & 1.4 & -- & 177.3 \\
TRIPLE\_PRUNER (sync) & Retrieval & 5.1 & 1.4 & -- & 25.0 \\
IR-CoT step (reasoning) & Reasoning & -- & -- & 8.3 & 121.8 \\
IR-CoT final answer & Reasoning & -- & -- & 1.0 & 223.7 \\
\midrule
\textbf{Total} & -- & \textbf{204.1} & \textbf{58.0} & \textbf{9.3} & -- \\
\textbf{Output tok/q} & -- & \textbf{27.0K} & \textbf{7.8K} & \textbf{1.2K} & -- \\
\bottomrule
\end{tabular}
}
\caption{Bamboogle inference cost by LLM role. Call counts are averaged per question; the role-level token column reports tokens per call, and the final row reports total output tokens per question.}
\label{tab:cost_per_role}
\end{table*}

\section{Failure Mode Analysis}
\label{sec:appendix_failures}

Table~\ref{tab:failure_modes} reports the full failure breakdown for Co-E (MCTS) on Bamboogle: 29 incorrect predictions out of 125 questions. Categories are assigned by manual inspection of the prediction, working memory, and retrieved evidence. \textit{Retrieval gap} denotes cases where the system identifies the right entity but the corpus supports an incorrect fact. \textit{Incomplete chain} covers failures where the needed evidence is present but the final chain is too short, too long, or off by one hop. \textit{Wrong entity linking} covers scope-sensitive mentions resolved to the wrong entity. \textit{Synthesis error} means the correct answer appears in memory but the final generator outputs a different answer. \textit{Surface-form mismatch} captures strict-EM artifacts where the prediction is semantically correct but formatted differently from the gold string.

\begin{table*}[h]
\centering
\resizebox{\textwidth}{!}{
\begin{tabular}{lcl}
\toprule
Category & \% & Representative example \\
\midrule
Retrieval gap         & 44.8 & "Oregon Trail/MECC (1971)" is the longest-running franchise; corpus attributed it to Super Mario/Nintendo \\
Incomplete chain      & 20.7 & "Father of father of computer science" $\rightarrow$ Rev. John Robert Turing (correct: Julius Mathison Turing) \\
Wrong entity linking  & 13.8 & "Third oldest surviving university" $\rightarrow$ Durham (correct: Cambridge; Durham is third in England only) \\
Synthesis error       & 13.8 & "Habiba Akumu Nyanjango" in working memory; synthesizer output "Habiba Akumu Obama" \\
Surface-form mismatch &  6.9 & "2011" predicted (gold: "30 June 2011") \\
\bottomrule
\end{tabular}
}
\caption{Failure mode breakdown for Co-E (MCTS) on Bamboogle.}
\label{tab:failure_modes}
\end{table*}

\section{Prompt Templates}
\label{sec:appendix_prompts}

This appendix reproduces the prompt templates used at each stage of Co-E. All LLM calls use JSON-mode structured output; each prompt's \texttt{Output Format} block gives the required schema. Prompts are shared across KGQA and text-QA benchmarks, with only runtime context variables changed (question, memory state, and retrieved evidence).

\subsection{Request Generation}
Used at the start of each reasoning step (\S~\ref{sec:overview}, stage 1) to generate a focused sub-query $q^{(t)}$ from the current memory $\mathcal{M}^{(t)}$ and question $q$. In CoT mode, this prompt also decides whether the question is already answerable from memory. In MCTS mode, it serves as the \textbf{Sub-question/Answer} node type.

\begin{promptbox}{Request Generation Prompt}
\begin{lstlisting}[style=prompt]
You are an expert assistant for multi-hop question answering and reasoning
decomposition. Decide whether the main question can already be answered from the
provided context. If not, generate strategic subquestions to advance the reasoning.

## Intermediate Answer
If `intermediate_answer` is provided, it is the resolved result of the PREVIOUS
hop. Use it as the anchor for the next subquestion, do NOT re-ask what was
already resolved.

## Core Principles
Each subquestion must:
- target a real knowledge gap not answerable from the provided context
- be atomic, self-contained, and understandable without the main question
- be non-redundant with other generated subquestions

**Sequential chains are allowed** when the main question explicitly links hops. In
these cases, generate subquestions in order, the first hop first, the next
anchored to its result. Sequential subquestions may depend on each other.

## Instructions
1. Analyze the main question: identify core intent, key entities, constraints,
   and required reasoning steps.
2. Check the context: if sufficient to answer, set `is_answerable` to true
   and stop.
3. Identify missing knowledge: only gaps that meaningfully advance reasoning
   toward the answer.
4. Generate subquestions:
   - For parallel gaps: each must be independently answerable.
   - For chained hops: generate in sequential order; use `intermediate_answer`
     to anchor step 2+.
   - If a subquestion is answerable with high confidence from common knowledge,
     include the answer inline.
5. Validate: remove subquestions answerable from context, redundant, or
   low-value. Keep at most 3.

## Scope Consistency
Preserve the geographic or categorical scope of the main question. Do NOT
silently narrow a global scope to a specific region without justification.

## Output Format
Respond with a JSON object with exactly these keys:
- is_answerable: boolean
- subquestions: array of strings or null
\end{lstlisting}
\end{promptbox}

\subsection{Corpus Search}

\subsubsection{Query Generator}
Rewrites the sub-query $q^{(t)}$ into search-optimized queries for web or dense-retrieval search (\S~\ref{sec:retrieval}, Corpus Search).

\begin{promptbox}{Query Generator Prompt}
\begin{lstlisting}[style=prompt]
You are a Reasoning Engine that deconstructs user input into precise,
self-contained search queries.

## Principles
1. Self-Contained: Each query understandable without original input.
2. Atomic: One single fact per query.
3. Essential & Non-Redundant: Every query necessary and unique.

## Instructions
1. Parse the Input: identify type (factual, comparative, causal, temporal),
   key entities, and required reasoning steps.
2. Generate Strategic Queries: formulate queries that collectively cover all
   necessary information to answer the input.
3. Ensure Self-Containment: each query must be independently answerable.
4. Review for Completeness and Non-Redundancy.
5. Temporal Grounding: if the input contains "current", "now", or a
   present-tense superlative, add "as of [current year]" to at least one query.
6. Fallback Queries: add 1-2 fallback queries using common aliases or
   alternative phrasings so that retrieval succeeds if the primary phrasing
   returns no results.

## Output Format
Respond with a JSON object with exactly these keys:
- queries: array of strings
\end{lstlisting}
\end{promptbox}

\subsubsection{Extractor}
Filters each retrieved document into self-contained, question-relevant snippets that form $S^{(t)}$ (\S~\ref{sec:retrieval}, Corpus Search). Raw documents never enter the working memory directly.

\begin{promptbox}{Extractor Prompt}
\begin{lstlisting}[style=prompt]
You are a meticulous research analyst. Build a comprehensive dossier of
information from the provided text that could help answer the question.

Rules:
- Consider both direct and indirect relevant information. Information is
  relevant if it contains clues that could help answer the question (not
  necessarily directly answering it, but providing information that could
  help answer the question).
- Extracted information must be self-contained and clear, i.e., understandable
  without any external context.

Instructions:
1. Question Deconstruction: identify primary subject, key entities, and
   specific information sought.
2. Candidate Identification: identify and quote ALL passages potentially
   related to concepts in the question. Be liberal and inclusive in this
   initial pass.
3. Relevance Evaluation: assess each quote against criteria (directly
   answering, contextual, supporting evidence, etc.).
4. Extraction: extract ALL relevant information. Add context for clarity but
   preserve original meaning. Each extracted item must be FULLY UNDERSTANDABLE
   on its own.
5. Final evaluation: examine each item for self-containment and relevance;
   rewrite or remove as needed.

## Output Format
Respond with a JSON object with exactly these keys:
- relevant_information: array of strings
\end{lstlisting}
\end{promptbox}

\subsection{Response Generation}
Used in stage 3 of each reasoning iteration (\S~\ref{sec:overview}) to produce an intermediate answer from the current memory augmented with newly retrieved evidence.

\begin{promptbox}{Response Generation Prompt}
\begin{lstlisting}[style=prompt]
You are an expert assistant specializing in precise, well-reasoned question
answering. Deliver a direct, accurate answer with transparent, step-by-step
reasoning.

## Instructions
1. Analyze the question: identify core intent, key entities, and specific
   information sought.
2. Context priority: when context is provided, ground your answer exclusively
   in the context, do not introduce external facts. Only use your own
   knowledge when context is absent or clearly incomplete, and explicitly state
   when doing so.
3. Synthesize a clear, well-reasoned answer. State any assumptions clearly.

## Output Format
Respond with a JSON object with exactly these keys:
- answer: string
- concise_answer: string
- reasoning: string
- confidence_level: string
\end{lstlisting}
\end{promptbox}

\subsection{Memory Updating}
\subsubsection{Memory Consolidation}
Runs twice per synchronization cycle (\S~\ref{sec:sync}): once after evidence retrieval (primary consolidation) and once after graph-to-text injection (re-consolidation).

\begin{promptbox}{Memory Consolidation Prompt}
\begin{lstlisting}[style=prompt]
You are an expert Memory Consolidation Agent. Process an input memory (a list
of information items) and consolidate it into a refined memory containing only
information relevant and useful for answering the given question.

## Instructions
1. Question Analysis: identify primary subject, key entities, and information
   sought.
2. Memory Atomization: atomize memory into atomic, self-contained items. No
   pronouns or references to external context.
3. Deduplication: if two items have the same content, keep one. If one item
   is completely contained in another, remove it.
4. Relevance Evaluation: keep items that contain ANY clue that could help
   answer the question.
4b. Provenance Audit: for each [System Prediction] item, check if any
    [Retrieval] item covers the same claim. If contradicted -> remove the
    [System Prediction] item. If supported -> upgrade provenance to "Retrieval".
4c. Hop Depth Filtering: items tagged [hop=N] were retrieved at reasoning
    step N. If a lower-hop item fully answers the question, discard higher-hop
    items not needed. Strip [hop=N] prefixes from output `content`; record
    depth in `hop_depth`.
5. Irrelevant Information Removal.
6. Conflict Resolution: when a [Retrieval] item and a [System Prediction] item
   state conflicting specific facts, ALWAYS keep [Retrieval] and discard
   [System Prediction]. If two [Retrieval] items conflict, keep both and note
   the conflict.
7. Refinement: ensure each item is self-contained and clear.
8. Final check: verify every kept item is self-contained, non-redundant, and
   has correct provenance.

## Output Format
Respond with a JSON object with exactly these keys:
- consolidated_memory: array of objects; each object has:
  - content: string
  - provenance: string, exactly "System Prediction" or "Retrieval"
  - hop_depth: integer or null
\end{lstlisting}
\end{promptbox}

\subsubsection{Named Entity Recognition and Entity Linking}
Identifies entities in the consolidated memory and links them to Wikidata QIDs for inclusion in the shared entity dictionary (\S~\ref{sec:sync}, text-to-graph propagation).

\begin{promptbox}{NER and Entity Linking Prompt}
\begin{lstlisting}[style=prompt]
You are an expert Named Entity Recognition specialist. Extract all named
entities from the text.

You may be given an optional list of KNOWN ENTITIES, each with:
- id: Wikidata QID
- name: official Wikidata label
- description: brief description for disambiguation

Wikidata linking rules (strict):
- If an extracted entity clearly matches a KNOWN ENTITY, set its id to that
  QID and use the KNOWN ENTITY's official name.
- If there is any ambiguity or you are not fully certain, set id to null.
- NEVER guess or invent a QID if it is not provided.

Instructions:
1. If text is a question, focus ONLY on entities which are main clues to
   answer the question.
2. Define precise boundaries (include modifiers).
3. Handle ambiguity using context.
4. Extract unique entities only once (deduplicate by real-world identity).

## Output Format
Respond with a JSON object with exactly these keys:
- entities: array of objects; each object has:
  - id: string or null
  - name: string
  - description: string or null
\end{lstlisting}
\end{promptbox}

\subsubsection{Relation Extraction}
Extracts open-vocabulary (subject, relation, object) triples from the consolidated memory to extend the graph memory $\mathcal{G}$ (\S~\ref{sec:sync}, text-to-graph propagation).

\begin{promptbox}{Relation Extraction Prompt}
\begin{lstlisting}[style=prompt]
You are an expert Relation Extraction specialist. Extract all meaningful
relationships between entities. Each relationship must be self-contained.

You may be given an optional list of KNOWN ENTITIES (with Wikidata QIDs);
apply the same strict linking rules as NER.

Instructions:
1. Identify entity pairs with direct relationships.
2. Break down complex relationships into simpler ones.
3. Only extract explicitly stated or strongly implied relationships.
4. Use clear, concise, active-voice relation types.
5. Ensure relations are self-contained and non-duplicated.

## Canonical Relation Direction
Always use the ACTIVE form. Never use passive or inverse forms

Rules, convert to active form when a predicate:
1. ends in "_of"  -> invert: has_child, contains, has_capital
2. starts with "is_" -> invert to active form
3. ends in "_by" -> invert: precedes, directed, owns, succeeds

## Output Format
Respond with a JSON object with exactly these keys:
- relations: array of objects; each object has:
  - subject: string
  - subject_id: string or null
  - relation: string
  - object: string
  - object_id: string or null
  - context: string or null
\end{lstlisting}
\end{promptbox}

\subsubsection{Triple Pruner}
Filters extracted and retrieved triples for question-relevance before they are merged into $\mathcal{G}$ (\S~\ref{sec:sync} and \S~\ref{sec:retrieval}). Applied as the LLM stage of the two-stage pruner (cross-encoder reranker followed by this LLM filter).

\begin{promptbox}{Triple Pruner Prompt}
\begin{lstlisting}[style=prompt]
You are a Knowledge Graph Expert. Given a question and a list of triples,
keep only the triples that are genuinely useful for answering the question.

## Relevance Criteria
A triple (Subject, Relation, Object) is relevant ONLY if it meets one of:

1. Direct relevance: both subject AND object are directly related to the
   question, and the relation connects them in a way that helps answer it.
   - Example Q: "What is the capital of France?"
     Keep: (France, capital, Paris)
     Drop: (France, borders, Germany)

2. Chain relevance: the triple forms part of a reasoning chain with another
   kept triple. One entity of this triple must match an entity in another
   relevant triple, and together they help answer the question.
   - Example Q: "Who is the spouse of the president of France?"
     Keep: (France, president, Macron) + (Macron, spouse, Brigitte)

## Key Rule
Do NOT keep a triple just because one entity superficially matches a word in
the question. DO keep a triple if one entity is clearly relevant AND the other
could plausibly be an intermediate step or answer in the reasoning chain.

## Output Format
Respond with a JSON object with exactly these keys:
- keep_indices: array of integers, 0-based indices of triples to retain
\end{lstlisting}
\end{promptbox}

\subsection{MCTS-Specific Prompts}

These prompts are used exclusively in MCTS mode (\S~\ref{sec:modes}). Each corresponds to one of the five MCTS node types.

\subsubsection{Self-Correction Node}

\begin{promptbox}{Self-Correction Prompt}
\begin{lstlisting}[style=prompt]
You are an expert in answer verification and refinement. Given a question,
proposed answer, and context, verify correctness and provide a refined response.

## Instructions
1. Parse question requirements.
2. Extract relevant facts from context.
3. Evaluate proposed answer as: CORRECT / PARTIAL / INCORRECT / UNSUPPORTED.
4. Generate refined answer.

## Output Format
Respond with a JSON object with exactly these keys:
- status: string, one of: correct, partial, incorrect, unsupported
- refined_answer: string
- confidence_level: string
\end{lstlisting}
\end{promptbox}

\subsubsection{Question Rephrasing Node}

\begin{promptbox}{Question Rephrasing Prompt}
\begin{lstlisting}[style=prompt]
You are a Question Refiner that transforms unclear questions into precise,
clear questions. The rephrased question must be fully understandable on its
own without needing to refer back to the original.

## Principles
1. Clarity First: eliminate ambiguity and jargon.
2. Preserve Intent: do not alter the core inquiry.
3. Enhance Answerability: make specific and self-contained.

## Output Format
Respond with a JSON object with exactly these keys:
- rephrased_question: string
\end{lstlisting}
\end{promptbox}

\subsubsection{Synthesis Node}

\begin{promptbox}{Synthesis Prompt}
\begin{lstlisting}[style=prompt]
You are a specialized AI for multi-step reasoning. Perform a single, focused
reasoning step by analyzing context and producing a consolidated synthesis.

## Instructions
1. Analyze the main question objective.
2. Review all information in context.
3. If context is sufficient to directly answer, state this and formulate the
   definitive answer.
4. Otherwise, synthesize new thoughts that advance reasoning:
   - Causal or temporal links
   - Core relationship identification
   - Progress summary
   - Contradiction identification
   - Hypothesis formulation

## Critical Constraints
1. No External Information: do NOT introduce facts not in the context.
2. No New Questions: synthesize, do not query.

## Output Format
Respond with a JSON object with exactly these keys:
- is_answerable: boolean
- step_conclusion: string
- confidence_level: string
\end{lstlisting}
\end{promptbox}

\subsubsection{Final Answer Synthesis}
Synthesizes the final answer from all candidate terminal nodes, weighted by their MCTS reward scores.

\begin{promptbox}{Final Answer Synthesis Prompt}
\begin{lstlisting}[style=prompt]
You are an expert in argumentative synthesis. Construct a superior answer by
critically analyzing candidate answers and grounding your synthesis in
supporting evidence.

## Understanding your inputs
Candidate answers are MCTS outputs prefixed with [quality_score=X] where X in
[-1, 1]. Score >= 0.5: primary evidence. Score 0-0.5: supporting evidence.
Score < 0: treat skeptically.

supporting_evidence has two components:
- Textual facts tagged [Retrieval] (high reliability) or [System Prediction]
  (medium reliability).
- Knowledge graph triples (under **Information N** sections): structured
  entity-relationship facts. Use for verifying entity attributes; discard
  triples unrelated to the question.

## Synthesis Procedure
Phase I, Candidate Analysis: extract core claim, premises, reasoning chain
for each candidate. Weight by quality_score.

Phase II, Evidence Cross-Check & Conflict Resolution. Adjudication hierarchy:
1. High-score (>=0.5) + [Retrieval] corroboration
2. Convergent positive-score candidates
3. [Retrieval] facts regardless of candidate score
4. Logically sound reasoning consistent with graph triples
5. Majority agreement as last resort

Phase III, Synthesis & Self-Critique: verify every factual claim is traceable
to at least one reliable source. Revise if not.

## Output Format
Respond with a JSON object with exactly these keys:
- final_answer: string
- concise_answer: string
- reasoning: string
- confidence_level: string, one of: high, medium, low, uncertain
\end{lstlisting}
\end{promptbox}

\subsubsection{Verifier}
The verifier runs three parallel calls conditioned on (i) no retrieved context, (ii) the textual memory $\mathcal{T}$, and (iii) the graph memory $\mathcal{G}$ serialized as triples; the normalized mean score becomes the MCTS reward $r \in [-1, 1]$.

\begin{promptbox}{Verifier Prompt}
\begin{lstlisting}[style=prompt]
You are an expert verifier. Given a question and a candidate answer, evaluate
how well the answer is supported by the available evidence.

- If context is provided: use it as the primary source to verify the answer.
- If context is not provided: use your own knowledge to verify independently.

Rate from 0.0 to 10.0 how well the answer is correct/supported.

Scoring:
- 9.0-10.0: Fully verified / strongly supported
- 7.0-8.9:  Mostly supported, minor gaps
- 5.0-6.9:  Partially supported or uncertain
- 3.0-4.9:  Weakly supported, significant doubts
- 0.0-2.9:  Contradicted or completely unsupported

## Output Format
Respond with a JSON object with exactly these keys:
- rating: number, float from 0.0 to 10.0
- reasoning: string
\end{lstlisting}
\end{promptbox}

\subsection{LLM-Judged Accuracy (Acc)}
Used offline to compute the supplementary Acc metric. This judge is not part of the Co-E inference pipeline.

\begin{promptbox}{LLM-Judged Accuracy Prompt}
\begin{lstlisting}[style=prompt]
You are an expert evaluator. Rate the system_answer on a scale from 0.0 to
10.0 based on how effectively it addresses the user_question.

Evaluation Criteria:
1. Correctness (60%): Is the information factually accurate?
   - If correct_answer is provided and system_answer matches, award 10.0.
   - Otherwise, verify accuracy using your knowledge.
2. Helpfulness & Relevance (40%): Does it address the user's core need?

Scoring Guidelines:
- 9.0-10.0: Correct and comprehensive
- 7.0-8.9:  Mostly correct with minor issues
- 5.0-6.9:  Partially addresses or has accuracy concerns
- 3.0-4.9:  Significant correctness or relevance issues
- 0.0-2.9:  Incorrect or completely off-topic

Uncertainty rule: if you are not confident about the correct answer (obscure
facts, precise numbers, rare entities), assign 5.0 rather than a confident
high or low score.

## Output Format
Respond with a JSON object with exactly these keys:
- rating: number, float from 0.0 to 10.0
- reasoning: string
\end{lstlisting}
\end{promptbox}

\end{document}